%% file: main.tex
\documentclass[10pt,twocolumn,letterpaper]{article}

\usepackage{cvpr}              

\usepackage[accsupp]{axessibility}
\usepackage{algorithm}
\usepackage{algpseudocode}

\usepackage{amsmath,amsfonts}
\usepackage{booktabs}
\usepackage{multirow}
\usepackage[table,xcdraw,dvipsnames]{xcolor}
\usepackage{array}
\usepackage{graphicx}
\usepackage{subcaption}
\newcolumntype{C}[1]{>{\centering\arraybackslash}p{#1}}

\definecolor{cvprblue}{rgb}{0.21,0.49,0.74}
\usepackage[pagebackref,breaklinks,colorlinks,allcolors=cvprblue]{hyperref}

\def\paperID{*****} 
\def\confName{CVPR}
\def\confYear{2026}

\title{CausalVAD: De-confounding End-to-End Autonomous Driving\\via Causal Intervention}

\author{
Jiacheng Tang$^{1}$\thanks{These authors contributed equally to this work.} \quad
Zhiyuan Zhou$^{1}$\footnotemark[1] \quad
Zhuolin He$^{1}$ \quad
Jia Zhang$^{3}$ \quad
Kai Zhang$^{4}$ \quad
Jian Pu$^{1, 2}$\thanks{Corresponding author.} \\
$^{1}$ Fudan University \quad
$^{2}$ Embodiq Robotics Co., Ltd. \\
$^{3}$ Beijing Institute of Technology \quad
$^{4}$ East China Normal University \\
{\tt\small \{jiachengtang21, zhouzy25, zlhe22\}@m.fudan.edu.cn,} \\
{\tt\small zhangjia@bit.edu.cn, kzhang@cs.ecnu.edu.cn, jianpu@fudan.edu.cn} \\
}

\begin{document}
\maketitle

\begin{abstract}

Planning-oriented end-to-end driving models show great promise, yet they fundamentally learn statistical correlations instead of true causal relationships. This vulnerability leads to causal confusion, where models exploit dataset biases as shortcuts, critically harming their reliability and safety in complex scenarios. To address this, we introduce CausalVAD, a de-confounding training framework that leverages causal intervention. At its core, we design the sparse causal intervention scheme (SCIS), a lightweight, plug-and-play module to instantiate the backdoor adjustment theory in neural networks. SCIS constructs a dictionary of prototypes representing latent driving contexts. It then uses this dictionary to intervene on the model's sparse vectorized queries. This step actively eliminates spurious associations induced by confounders, thereby eliminating spurious factors from the representations for downstream tasks. Extensive experiments on benchmarks like nuScenes show CausalVAD achieves state-of-the-art planning accuracy and safety. Furthermore, our method demonstrates superior robustness against both data bias and noisy scenarios configured to induce causal confusion.

\end{abstract}

\section{Introduction}
\label{sec:introduction}

In recent years, planning-oriented end-to-end autonomous driving models have achieved remarkable progress~\cite{hu2023planning, jiang2023vad, sun2025sparsedrive, chen2024ppad}. By jointly optimizing perception, prediction, and planning within a unified framework, these models attempt to directly map raw sensor data to final driving decisions. Seminal works like UniAD~\cite{hu2023planning} and VAD~\cite{jiang2023vad} have enhanced model interpretability by exposing structured intermediate representations of the scene (e.g., object trajectories, map elements), partially opening up the black box. Nevertheless, a more profound challenge emerges: \textit{how to ensure the explanations provided by the model are both logically valid and faithful to its internal decision-making mechanism}~\cite{atakishiyev2025safety, jiang2025survey, xie2025vlms}?

\begin{figure}[t]
    \centering
    \includegraphics[width=.92\linewidth]{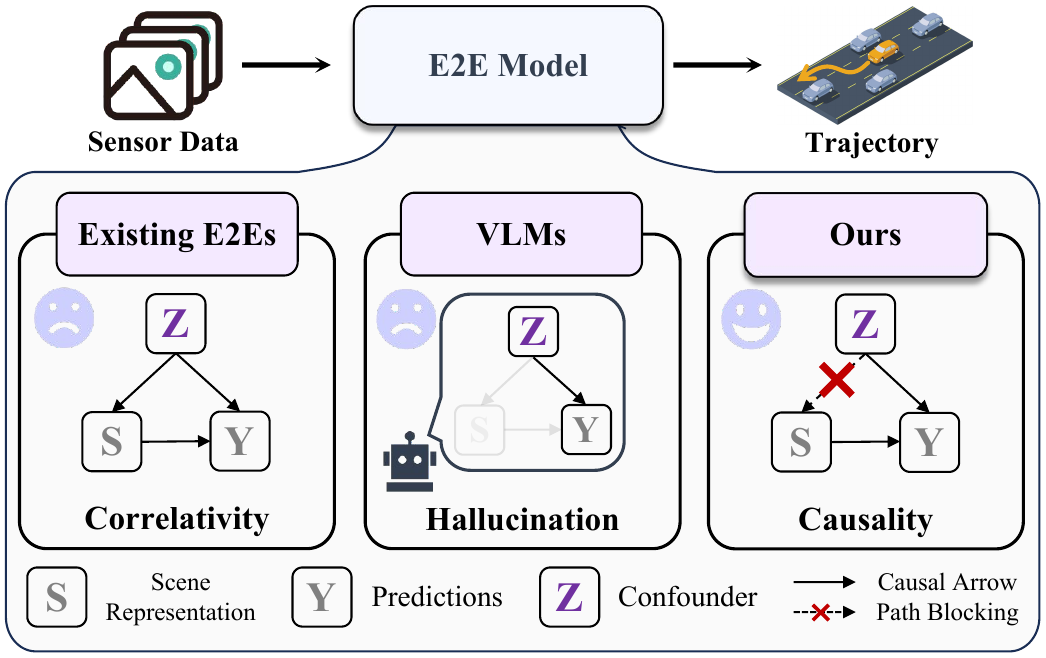}
    \caption{
        The problem of spurious correlation.
        (Left) Standard end-to-end models learn the observational correlation $P(Y|S)$, making them vulnerable to latent confounders $Z$ (e.g., scene context) that create a spurious backdoor path ($S \leftarrow Z \rightarrow Y$).
        (Middle) VLM-based approaches suffer from the same confounding and introduce hallucinations.
        (Right) Our CausalVAD performs a causal intervention $P(Y|\text{do}(S))$ via backdoor adjustment, severing the spurious link to learn the true causal effect for robust and trustworthy decision-making.
    }
    \label{fig:interpretability}
\end{figure}

Current efforts to improve the decision interpretability of these models primarily follow two paths. The first path, refining the multi-task learning paradigm, essentially learns the statistical correlation among modules rather than the underlying causal relationships, as depicted in \Cref{fig:interpretability} (left). This renders models prone to exploiting spurious correlations in the dataset, a phenomenon known as \textit{shortcut learning}~\cite{geirhos2020shortcut, li2024ego}. For instance, in an imitation learning setup, a model might discover that the ego-vehicle's own past kinematic states (e.g., velocity, acceleration) are highly predictive of the expert's future decisions. This reliance on its own history, a form of spurious self-correlation, performs exceptionally well in open-loop evaluations~\cite{li2024ego}. In real-world closed-loop deployment, however, any minor deviation from the expert's trajectory gets rapidly amplified through this historical dependency, leading to catastrophic performance degradation~\cite{li2024ego, cheng2024rethinking}. This phenomenon is a classic manifestation of \textit{causal confusion}~\cite{muller2005off, de2019causal, tang2025decoupling}. The model mistakenly identifies a correlated variable, its own past state, as the cause for its next maneuver. In reality, the true driver for its behavior should be a reaction to the surrounding environment. The other mainstream approach involves integrating large vision-language models (VLMs) to endow models with higher-level reasoning~\cite{tian2024drivevlm, xu2025drivegpt4, yan2025ordermind}. VLMs can leverage their vast world knowledge to generate natural language explanations for driving decisions. Their inherent problems, such as \textit{hallucination}~\cite{liu2024survey, jiang2025survey} and \textit{pseudo-faithfulness}~\cite{xie2025vlms}, severely limit their application in the safety-critical domain of autonomous driving. As shown in \Cref{fig:interpretability} (middle), VLM-generated explanations may sound plausible but can be entirely disconnected from the model's actual decision-making process~\cite{xie2025vlms},  introducing new and unpredictable risks to system reliability.

For the pursuit of trustworthy autonomous driving, we contend that layering an unreliable explanation module onto a flawed correlation-based learning framework may not be a viable path. Instead, we advocate for a fundamental reshaping of the model's learning objective: shifting from fitting correlations to inferring causation. We posit that a modular, multi-task architecture is the foundation for reliable interpretability, but only if the connections between its modules are causal. To this end, we propose a novel training paradigm that aims to learn the true causal effect from scene representation to planning, i.e., $P(\text{planning} | \text{do}(\text{scene}))$~\cite{pearl2009causality}, rather than the conventional conditional probability $P(\text{planning} | \text{scene})$, as illustrated in \Cref{fig:interpretability} (right). Specifically, we first formulate a structural causal model (SCM) for end-to-end autonomous driving and identify the backdoor path introduced by latent confounders (e.g., scene context) that leads to spurious correlations. Guided by the principle of backdoor adjustment from causal inference~\cite{pearl2009causality}, we design a sparse causal intervention scheme (SCIS). SCIS operationalizes the do-operator within the neural network by leveraging a pre-computed confounder dictionary to represent latent driving contexts and performing a parameterized causal intervention on the efficient vectorized scene queries of the VAD model. This process severs the formation of spurious associations, compelling the planning module to make decisions based on the true causal effects of the scene representation. We instantiate SCIS on the VAD architecture, proposing CausalVAD.

We conduct extensive experiments on multiple benchmarks, including nuScenes~\cite{caesar2020nuscenes}, NAVSIM~\cite{dauner2024navsim}, and Bench2Drive~\cite{jia2024bench2drive}. The results demonstrate that CausalVAD not only surpasses state-of-the-art methods on standard evaluation metrics, but crucially exhibits significantly enhanced robustness in noisy scenarios designed to induce causal confusion. Our contributions are as follows:
\begin{itemize}
    \item We introduce a formal causal analysis of modern planning-oriented end-to-end autonomous driving models and propose a practical de-confounding framework.
    \item We propose SCIS, a plug-and-play design that parameterizes backdoor adjustment. We reveal and leverage the intrinsic synergy between VAD's sparse vectorized representation and our causal intervention mechanism.
    \item Our CausalVAD significantly improves planning safety and robustness across multiple datasets, offering a new pathway toward more trustworthy autonomous driving.
\end{itemize}

\section{Related work}
\label{sec:relatedwork}

\begin{figure*}[t]
    \centering
    \includegraphics[width=.92\textwidth]{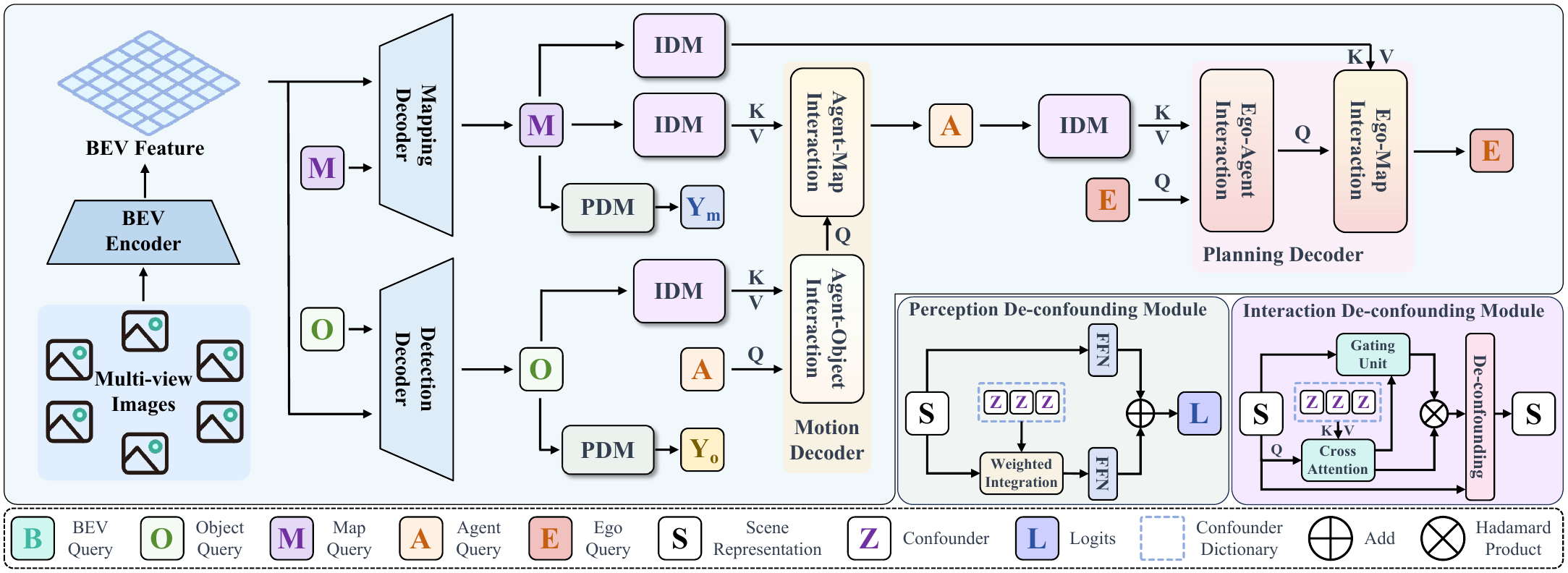}
    \caption{
        The overall architecture of CausalVAD. Our method performs precise, multi-stage causal interventions at critical information hubs within the VAD pipeline.
        (1) Perception stage (bottom left): The perception de-confounding module (PDM) operates on the classification logits ($Y_o, Y_m$). A dual-branch structure adjusts the direct classification score $L$ against a bias score derived from the confounder dictionaries ($\{\mathcal{Z}_o\}, \{\mathcal{Z}_m\}$), outputting de-confounded logits.
        (2) Prediction and planning stages (bottom right): The interaction de-confounding module (IDM) removes spurious factors from queries before fusion in downstream tasks. IDM utilizes cross-attention to estimate the spurious component predictable from the context. This component is scaled by a gating unit and subtracted from the original query to block spurious associations.
    }
    \label{fig:causalvad}
\end{figure*}

\paragraph{Planning-oriented end-to-end autonomous driving.}
Recent advances in end-to-end autonomous driving have largely focused on architectural innovations for enhanced scene representation. Early works, such as LBC~\cite{chen2020learning}, utilized implicit Bird's-Eye-View (BEV) representations. The paradigm then evolved to incorporate explicit and interpretable intermediate representations~\cite{yan2024pointssc, liartdeco}. One line of work employs rasterized BEV feature maps as a unified scene representation, exemplified by UniAD~\cite{hu2023planning} and ST-P3~\cite{hu2022st}. To further improve efficiency and precision, another line of work has shifted towards vectorized representations, which directly model scene elements like lanes and obstacles on sparse, learnable queries~\cite{li2025papl}. VAD~\cite{jiang2023vad} and SparseDrive~\cite{sun2025sparsedrive} are prominent examples of this approach. While these works have achieved great success at the network architecture level, they commonly adopt standard supervised learning objectives, which constrains their ability to learn causal relationships between modules. Our work is orthogonal to these architectural explorations. We focus on introducing a new paradigm aimed at fundamentally enhancing the model's causal reasoning capabilities.

\paragraph{Causal confounding in driving models.}
Causal confusion is a recognized challenge in data-driven driving models~\cite{muller2005off, de2019causal}, with shortcut learning being a typical manifestation~\cite{geirhos2020shortcut}. One line of work mitigates this issue using heuristic methods. For instance, studies have employed state dropout~\cite{cheng2024rethinking} or diverse data augmentation techniques~\cite{li2024ego} to force the model to reduce its reliance on spurious correlations. A more principled line of work attempts to apply causal theory. Some studies leverage counterfactual reasoning~\cite{wang2025omnidrive} or causal discovery algorithms~\cite{pourkeshavarz2024cadet} for offline analysis of driving behavior or for decision-making in simplified scenarios. However, seamlessly and efficiently integrating rigorous causal intervention theory into the online training of large-scale end-to-end models remains a significant challenge. Our work aims to fill this gap by proposing the scalable, plug-and-play, online de-confounding training scheme grounded in the theory of backdoor adjustment~\cite{muller2005off}.

\paragraph{VLM-based driving models.}
Large Vision-Language Models (VLMs) are being actively explored in autonomous driving for their powerful commonsense reasoning and interactive capabilities. Current applications can be broadly categorized into two types. The first category uses VLMs as a high-level decision brain for parsing complex scenes, predicting intentions, and making macroscopic plans~\cite{tian2024drivevlm, jiang2024senna}. The second category leverages VLMs to provide natural language explanations for the model's decisions, aiming to enhance system transparency~\cite{hwang2024emma, fu2025orion}. Despite their promise, a core challenge for these methods is the lack of guaranteed faithfulness~\cite{xie2025vlms}. The VLM's reasoning process may be disconnected from the underlying generation of the driving policy. This poses a significant risk in safety-critical applications. CausalVAD takes a different path. Instead of relying on external knowledge sources, we enforce the causal validity and internal consistency of the model's decision logic through causal intervention.

\section{Method}
\label{sec:method}

As illustrated in \Cref{fig:causalvad}, we propose CausalVAD, a de-confounding framework for mitigating spurious correlations in planning-oriented autonomous driving models. To address this, we design a novel SCIS by instantiating causal inference theory~\cite{pearl2009causality}. We detail its theoretical foundations, describe the construction of a multi-modal confounder dictionary, and present its two key causal intervention modules.


\subsection{Causal graph and problem formulation}
\label{subsec:causal_formalism}

\begin{figure}[t]
    \centering
    \begin{subfigure}{.273\linewidth}
        \centering
        \includegraphics[width=\linewidth]{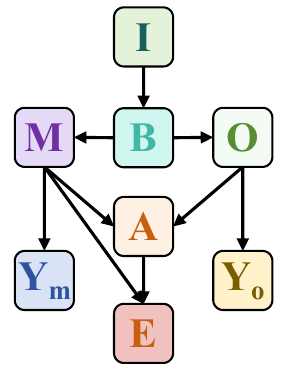}
        \caption{Causal graph}
        \label{fig:vad_totel_scm}
    \end{subfigure}
    \hfill
    \begin{subfigure}{.191\linewidth}
        \centering
        \includegraphics[width=\linewidth]{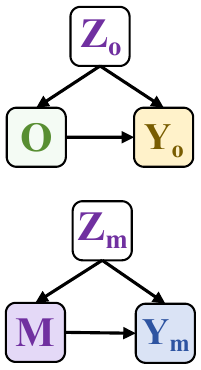}
        \caption{Perception}
        \label{fig:vad_p1_scm}
    \end{subfigure}
    \hfill
    \begin{subfigure}{.191\linewidth}
        \centering
        \includegraphics[width=\linewidth]{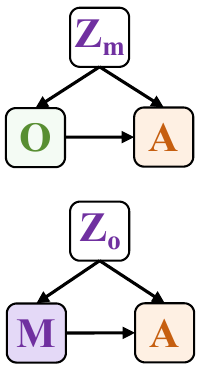}
        \caption{Prediction}
        \label{fig:vad_p2_scm}
    \end{subfigure}
    \hfill
    \begin{subfigure}{.191\linewidth}
        \centering
        \includegraphics[width=\linewidth]{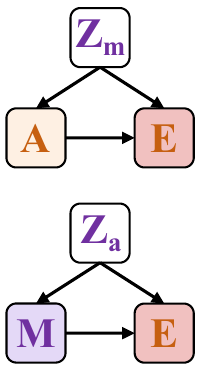}
        \caption{Planning}
        \label{fig:vad_p3_scm}
    \end{subfigure}
    \caption{The structural causal model (SCM) of VAD. The sub-figure (a) illustrates the causal flow between key representations. The sub-figures (b)-(d) on the right highlights several key confounders $Z$ that introduce backdoor paths.}
    \label{fig:vad_scm}
\end{figure}

To precisely locate and resolve issues arising from spurious correlations, we begin by formalizing VAD's modular pipeline as a structural causal model (SCM), depicted in \Cref{fig:vad_scm}. In this causal graph, nodes represent key intermediate representations, and directed edges denote the causal direction of information flow. Specifically, $\mathcal{I}$ represents multi-view image inputs, $\mathcal{B}$ is the unified BEV feature map, $\mathcal{O}$, $\mathcal{M}$, $\mathcal{A}$, $\mathcal{E}$ are the sparse queries for Objects, Map, Agents, and the Ego-vehicle, respectively, and $\mathcal{Y}_o$, $\mathcal{Y}_m$ are the classification outputs for the perception tasks.

By applying the d-separation criterion~\cite{pearl2009causality} to this graph, we identify three primary sources of causal confounding.

\textbf{Co-occurrence bias in perception.}
In the path $\mathcal{O} \rightarrow \mathcal{Y}_o$, the model aims to classify objects based on their query features. However, training data is replete with co-occurrence bias. For instance, \textit{pedestrians} and \textit{strollers} often appear together. This co-occurrence acts as a latent confounder $\mathcal{Z}_o$ (e.g., \textit{stroller is present in the scene}), opening a spurious path $\mathcal{O} \leftarrow \mathcal{Z}_o \rightarrow \mathcal{Y}_o$. This causes the model to erroneously use the \textit{stroller} context as a shortcut when classifying \textit{pedestrian}, rather than relying on the pedestrian's intrinsic visual features. Similarly, for the map task path $\mathcal{M} \rightarrow \mathcal{Y}_m$, analogous co-occurrence biases (e.g., \textit{traffic light} and \textit{crosswalk}) introduce a confounder $\mathcal{Z}_m$, opening a backdoor path $\mathcal{M} \leftarrow \mathcal{Z}_m \rightarrow \mathcal{Y}_m$ and degrading map element classification.

\textbf{BEV features as a common cause.}
In the motion prediction task, the generation of the Agent query $\mathcal{A}$ depends on both the Object query $\mathcal{O}$ and the Map query $\mathcal{M}$. Since both $\mathcal{O}$ and $\mathcal{M}$ are generated from BEV features $\mathcal{B}$, $\mathcal{B}$ acts as a common cause. This opens two symmetric backdoor paths: for $\mathcal{O} \rightarrow \mathcal{A}$, the path $\mathcal{O} \leftarrow \mathcal{B} \rightarrow \mathcal{M} \rightarrow \mathcal{A}$ exists, and for $\mathcal{M} \rightarrow \mathcal{A}$, the path $\mathcal{M} \leftarrow \mathcal{B} \rightarrow \mathcal{O} \rightarrow \mathcal{A}$ exists. This confounding makes it difficult to disentangle whether an agent's motion (from $\mathcal{A}$) is a reaction to dynamic objects (from $\mathcal{O}$) or to the static map (from $\mathcal{M}$), as both representations are spuriously correlated by their common origin $\mathcal{B}$.

\textbf{Input correlation in planning.}
In the planning task, the Ego query $\mathcal{E}$ bases its decision on the dynamic Agents $\mathcal{A}$ and the static Map $\mathcal{M}$. These two inputs, however, are highly correlated, as they share a deep common cause $\mathcal{B}$ and $\mathcal{M}$ is a cause of $\mathcal{A}$. This correlation confounds the planner, making it difficult to learn the pure causal contribution of the $\mathcal{A} \rightarrow \mathcal{E}$ and $\mathcal{M} \rightarrow \mathcal{E}$ paths independently. For instance, if \textit{agents braking} (from $\mathcal{A}$) occurs predominantly at \textit{intersections} (from $\mathcal{M}$) in the training data, the model may learn a spurious shortcut \textit{intersection} $\rightarrow$ \textit{decelerate}. This could cause the ego-vehicle to slow down unnecessarily at empty intersections or, conversely, react sluggishly to a braking agent in a non-intersection setting.

The above analysis reveals that causal confounding in the VAD framework is not a single failure point but a cascading systemic issue. Within VAD's information flow, perception, prediction, and planning are sequential stages. The classification bias in perception, representation confounding in prediction, and input correlation confounding in planning are thus distinct problems at different informational hubs. Therefore, a rigorous de-confounding solution must be multi-stage, applying precise, targeted causal interventions at each critical node.

\subsection{De-confounding via causal intervention}
\label{subsec:deconfounding}

\begin{figure}[t]
    \centering
    \includegraphics[width=.75\linewidth]{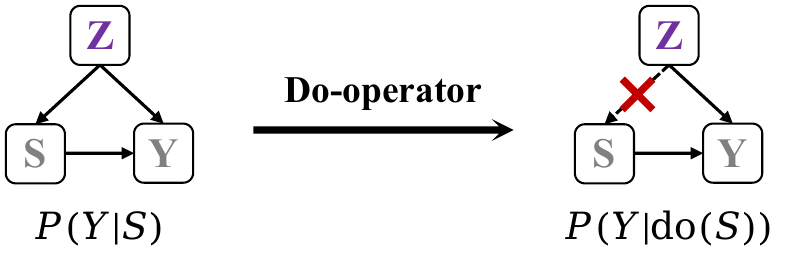}
    \caption{The backdoor adjustment \cite{pearl2009causality} principle. A confounder $Z$ opens a spurious backdoor path $S \leftarrow Z \rightarrow Y$. Applying the do-operator, i.e., $P(Y|\text{do}(S))$, severs this path, isolating the pure causal effect $S \rightarrow Y$.}
    \label{fig:backdoor_adjustment}
\end{figure}

Despite their different manifestations, all confounding issues identified above stem from a common structure in the causal graph: a backdoor path $S \leftarrow Z \rightarrow Y$ introduced by a confounder $Z$, which contaminates the target causal path $S \rightarrow Y$ (\Cref{fig:backdoor_adjustment}). Conventional supervised learning optimizes the observational probability $P(Y|S)$, which conflates the true causal effect with spurious associations.

Our key insight is that these backdoor paths can be actively severed by operationalizing Pearl's backdoor adjustment principle~\cite{pearl2009causality}. To this end, the learning objective should shift to estimating the post-intervention distribution $P(Y|\text{do}(S))$, which can be computed by:
\begin{equation}
    P(Y | \text{do}(S=s)) = \sum_z P(Y | S=s, Z=z) P(Z=z).
    \label{eq:backdoor_adjustment}
\end{equation}
The primary challenge in applying this formula is that the confounder $Z$ is latent and high-dimensional. We therefore propose SCIS, a general framework designed to tackle this challenge by operationalizing the backdoor adjustment theory. The core idea of SCIS is to model the latent confounder using a learnable proxy and to perform a parameterized intervention on the information flow within the neural network accordingly.

\subsection{Instantiation of SCIS}
\label{subsec:scis_instantiation}

We instantiate the SCIS framework into two key components: a multi-modal confounder dictionary and a set of parameterized causal intervention modules.

\subsubsection{Multi-modal confounder dictionary construction}

To model the latent confounder $Z$, we employ a proxy, a set of prototypes extracted from the data that are representative of different contexts. We construct this dictionary via a two-step offline process.

\textbf{Feature extraction.} We use a pre-trained VAD model to perform a single forward pass over the entire training set, collecting the sparse representations, i.e., the embeddings of Object queries $\mathcal{O}$, Map queries $\mathcal{M}$, and Agent queries $\mathcal{A}$.

\textbf{Prototype clustering.} We apply the clustering algorithm to the collected embeddings for each of the three query types separately. The centroid of each cluster is treated as a prototype for that class of information. All these prototypes collectively form our multi-modal confounder dictionary $\{\mathcal{Z}\} = \{\{\mathcal{Z}_o\}, \{\mathcal{Z}_m\}, \{\mathcal{Z}_a\}\}$.

\subsubsection{Parameterized causal intervention modules}

Based on the dictionary, we design two types of plug-and-play modules to address the different confounding problems identified. Detailed architectures of these modules are provided in the supplementary material.

\textbf{Perception de-confounding module (PDM).}
The PDM targets the \textit{co-occurrence bias in perception} within the classification paths $\mathcal{O} \rightarrow \mathcal{Y}_o$ and $\mathcal{M} \rightarrow \mathcal{Y}_m$. Its core mechanism is to adjust a query's classification logits by conditioning on its domain-specific prototype dictionary, thereby nullifying spurious contextual shortcuts. For object classification, the PDM takes a query $o \in \mathcal{O}$ and its dictionary $\{\mathcal{Z}_o\}$ to produce de-confounded logits:
\begin{equation}
    \text{logits}'_o = \text{PDM}(o, \{\mathcal{Z}_o\}).
\end{equation}
This intervention is applied symmetrically to the map classification path $\mathcal{M} \rightarrow \mathcal{Y}_m$ (using $\mathcal{Z}_m$), forcing the classifiers to focus on intrinsic query features.

\textbf{Interaction de-confounding module (IDM).}
The IDM resolves the more complex interaction confounding issues: \textit{BEV features as a common cause} and \textit{input correlation in planning}. It is a unified intervention architecture, instantiated multiple times for surgical interventions. Its principle is to refine an input query against a specific confounder proxy as dictated by our causal analysis.

We apply this module in a stage-wise de-confounding manner. To de-confound the prediction stage, we filter the inputs $\mathcal{O}$ and $\mathcal{M}$ to break their spurious correlation:
\begin{equation}
    \mathcal{O}' = \text{IDM}(\mathcal{O}, \{\mathcal{Z}_m\}) \text{,} \quad \mathcal{M}' = \text{IDM}(\mathcal{M}, \{\mathcal{Z}_o\}).
\end{equation}
Subsequently, to de-confound the planning stage, we decouple its highly correlated inputs $\mathcal{A}$ and $\mathcal{M}$:
\begin{equation}
    \mathcal{A}' = \text{IDM}(\mathcal{A}, \{\mathcal{Z}_m\}) \text{,} \quad \mathcal{M}'' = \text{IDM}(\mathcal{M}, \{\mathcal{Z}_a\}).
\end{equation}
This stage-wise de-confounding ensures that downstream modules learn from de-confounded representations.

\subsection{Training pipeline}
\label{subsec:training}

The training pipeline for CausalVAD consists of two stages. The first is the offline construction of the confounder dictionary $\{\mathcal{Z}\}$. This involves a single forward pass of a pre-trained VAD model over the training set to collect query embeddings, followed by K-means++ clustering to obtain the prototypes. This stage is fully unsupervised and performed only once. Afterwards, we insert the PDM and IDM modules, equipped with the dictionary $\{\mathcal{Z}\}$, into the VAD base architecture and train the entire CausalVAD network end-to-end from scratch. Training from scratch ensures the model learns de-confounded causal relationships from the outset, rather than correcting pre-existing spurious correlations. During training, the total loss function $\mathcal{L}_{\text{total}}$ remains identical to that of the original VAD. The full training procedure is detailed in Algorithm S.1 in the supplementary.

\section{Experiments}
\label{sec:experiments}

In this section, we conduct a comprehensive set of experiments to systematically evaluate our proposed CausalVAD. We aim to validate CausalVAD's performance on standard planning benchmarks and to deeply probe its advantages in learning robust causal relationships, particularly its generalization ability in the face of data bias and novel environments. Furthermore, we dissect the effectiveness of our core design, SCIS, through extensive ablation studies.

\subsection{Experimental setup}

\subsubsection{Datasets and metrics}

Our primary evaluation is conducted on the nuScenes~\cite{caesar2020nuscenes} dataset, a large-scale benchmark widely used for vision-based autonomous driving research. Its inherent imbalanced distribution of scenarios (e.g., ~75\% are straight-driving scenes) provides an ideal testbed for validating the effectiveness of our causal inference approach. We follow the official training/validation split and report on two key open-loop planning metrics: the average L2 displacement error (L2) to measure trajectory accuracy, and the collision rate (CR) to assess planning safety.

To further probe the model's generalization capabilities, our evaluation is extended to two additional benchmarks: (1) NAVSIM~\cite{dauner2024navsim}, a non-reactive simulation testbed, where we use its official PDM score (PDMS), and (2) Bench2Drive~\cite{jia2024bench2drive}, a closed-loop simulator, for which we adopt the official driving score (DS) and success rate (SR). Unless otherwise specified, all reported results are on the nuScenes validation set. Inference speed (FPS) is measured on a single NVIDIA RTX 3090 GPU.

\subsubsection{Implementation details}

Our CausalVAD is implemented using PyTorch and the MMDetection3D~\cite{mmdet3d2020} framework. The model configuration strictly follows VAD, predicting a 3-second trajectory from 2 seconds of history within a $60\text{m} \times 30\text{m}$ perception range. Our proposed SCIS is applied at critical information hubs within the perception, prediction, and planning stages of VAD to address the confounding issues that were identified. The sizes of the confounder dictionaries, i.e., the number of prototype contexts $(k_o, k_m, k_a)$, are set to $(10, 3, 6)$. We use the AdamW optimizer with an initial learning rate of $2 \times 10^{-4}$, a weight decay of $0.01$, and a CosineAnnealing scheduler. The model is trained for 60 epochs on 8 NVIDIA RTX 3090 GPUs with a batch size of 4 per GPU.

\subsection{Main results}

\subsubsection{Open-loop planning performance}

\begin{table}
    \caption{Open-loop planning performance on nuScenes.}
    \label{tab:planningperformance}
    \centering
    \small
    \setlength{\tabcolsep}{2.1pt} 
    \begin{tabular}{l|cccc|cccc|c}
        \specialrule{1.0pt}{0pt}{0pt}
        \multirow{2}{*}{\textbf{Method}} & \multicolumn{4}{c|}{\textbf{L2 (m) $\downarrow$}} & \multicolumn{4}{c|}{\textbf{Collision ($\%$) $\downarrow$}} & \multirow{2}{*}{\textbf{FPS}} \\
         & 1s & 2s & 3s & \cellcolor{gray!30} Avg. & 1s & 2s & 3s & \cellcolor{gray!30} Avg. &  \\
        \hline
        UniAD & 0.45 & 0.70 & 1.04 & \cellcolor{gray!30} 0.73 & 0.62 & 0.58 & 0.63 & \cellcolor{gray!30} 0.61 & 1.8 \\
        VAD-tiny $\S$ & 0.44 & 0.72 & 1.06 & \cellcolor{gray!30} 0.74 & 0.31 & 0.48 & 0.52 & \cellcolor{gray!30} 0.44 & 5.6 \\
        VAD $\S$ & 0.34 & 0.59 & 0.92 & \cellcolor{gray!30} 0.62 & 0.21 & 0.34 & 0.58 & \cellcolor{gray!30} 0.38 & 3.1 \\
        PPAD & 0.31 & 0.56 & 0.87 & \cellcolor{gray!30} 0.58 & 0.08 & 0.12 & 0.38 & \cellcolor{gray!30} 0.19 & 2.6 \\
        SparseDrive & 0.30 & 0.58 & 0.95 & \cellcolor{gray!30} 0.61 & 0.01 & 0.05 & 0.23 & \cellcolor{gray!30} 0.10 & 6.1 \\
        MomAD & 0.43 & 0.88 & 1.62 & \cellcolor{gray!30} 0.98 & 0.06 & 0.16 & 0.68 & \cellcolor{gray!30} 0.30 & 7.8 \\
        BridgeAD & 0.28 & 0.55 & 0.92 & \cellcolor{gray!30} 0.58 & \textbf{0.00} & \textbf{0.04} & \textbf{0.20} & \cellcolor{gray!30} \textbf{0.08} & 3.9 \\
        VeteranAD & 0.33 & 0.59 & 0.94 & \cellcolor{gray!30} 0.62 & 0.03 & 0.12 & 0.34 & \cellcolor{gray!30} 0.16 & 4.3 \\
        FusionAD & - & - & - & \cellcolor{gray!30} 1.03 & 0.25 & 0.13 & 0.25 & \cellcolor{gray!30} 0.21 & - \\
        \hline
        OmniDrive & 0.40 & 0.80 & 1.32 & \cellcolor{gray!30} 0.84 & 0.04 & 0.46 & 2.32 & \cellcolor{gray!30} 0.94 & - \\
        Senna & 0.37 & 0.54 & 0.86 & \cellcolor{gray!30} 0.59 & 0.09 & 0.12 & 0.33 & \cellcolor{gray!30} 0.18 & - \\
        ORION & 0.17 & 0.31 & 0.55 & \cellcolor{gray!30} 0.34 & 0.05 & 0.25 & 0.80 & \cellcolor{gray!30} 0.37 & - \\
        \hline
        Ours & \textbf{0.27} & \textbf{0.52} & \textbf{0.82} & \cellcolor{gray!30} \textbf{0.54} & 0.02 & 0.09 & 0.22 & \cellcolor{gray!30} 0.11 & 5.4 \\
        \specialrule{1.0pt}{0pt}{0pt}
        \multicolumn{10}{p{.96\linewidth}}{
            \textit{Note:} $\S$ denotes re-implement result.
        }
    \end{tabular}
\end{table}


We compare CausalVAD against a suite of state-of-the-art end-to-end and VLM-based autonomous driving methods on the nuScenes validation set. As shown in \Cref{tab:planningperformance}, CausalVAD achieves the lowest average L2 error (0.54m) among all methods, demonstrating superior trajectory planning accuracy. Compared to our baseline VAD-tiny, CausalVAD reduces the L2 error by 27$\%$ and the collision rate by a staggering 75$\%$ with almost no additional computational cost. This significant performance gain, especially the dramatic improvement in safety, strongly indicates that by introducing causal intervention, the model learns to better understand scene dynamics and make more reliable decisions. Even when compared to more powerful models like BridgeAD, our method shows a clear advantage in planning accuracy while maintaining a highly competitive collision rate. Detailed evaluation results for other tasks (e.g., perception and prediction) are provided in the supplementary.

\subsubsection{Analysis of causal robustness}

\begin{table}
    \caption{Scene generalization ability on nuScenes.}
    \label{tab:scenegeneralization}
    \centering
    \small
    \setlength{\tabcolsep}{2.5pt} 
    \begin{tabular}{l|c|cccc|cccc}
        \specialrule{1.0pt}{0pt}{0pt}
        \multirow{2}{*}{\textbf{Method}} & \multirow{2}{*}{\textbf{Nav.}} & \multicolumn{4}{c|}{\textbf{L2 (m) $\downarrow$}} & \multicolumn{4}{c}{\textbf{Collision ($\%$)} $\downarrow$} \\
         &  & 1s & 2s & 3s & \cellcolor{gray!30} Avg. & 1s & 2s & 3s & \cellcolor{gray!30} Avg. \\
        \hline
        \multirow{2}{*}{VAD-tiny} & ST & 0.45 & 0.73 & 1.07 & \cellcolor{gray!30} 0.75 & 0.35 & 0.53 & 0.57 & \cellcolor{gray!30} 0.48 \\
         & LR & 0.64 & 1.08 & 1.49 & \cellcolor{gray!30} 1.07 & 0.12 & 0.21 & 0.53 & \cellcolor{gray!30} 0.29 \\
        \hline
        \multirow{2}{*}{Ours} & ST & \textbf{0.28} & \textbf{0.53} & \textbf{0.83} & \cellcolor{gray!30} \textbf{0.55} & \textbf{0.03} & \textbf{0.09} & \textbf{0.21} & \cellcolor{gray!30} \textbf{0.11} \\
         & LR & \textbf{0.40} & \textbf{0.62} & \textbf{1.05} & \cellcolor{gray!30} \textbf{0.69} & \textbf{0.01} & \textbf{0.14} & \textbf{0.41} & \cellcolor{gray!30} \textbf{0.19} \\
        \specialrule{1.0pt}{0pt}{0pt}
        \multicolumn{10}{p{.96\linewidth}}{
            \textit{Note:} Navigation `ST' denotes go straight, while `LR' denotes turn left/right.
        }
    \end{tabular}
\end{table}

\begin{table}
    \caption{Reliance on ego status on nuScenes.}
    \label{tab:relianceonesonnus}
    \centering
    \small
    \setlength{\tabcolsep}{0.28pt} 
    \begin{tabular}{l|c|cccc|cccc}
        \specialrule{1.0pt}{0pt}{0pt}
        \multirow{2}{*}{\textbf{Method}} & \textbf{Velo.} & \multicolumn{4}{c|}{\textbf{L2 (m) $\downarrow$}} & \multicolumn{4}{c}{\textbf{Collision ($\%$) $\downarrow$}} \\
         & \textbf{Noise} & 1s & 2s & 3s & \cellcolor{gray!30} Avg. & 1s & 2s & 3s & \cellcolor{gray!30} Avg. \\
        \hline
        \multirow{5}{*}{VAD-tiny} & - & 0.44 & 0.72 & 1.06 & \cellcolor{gray!30} 0.74 & 0.31 & 0.48 & 0.52 & \cellcolor{gray!30} 0.44 \\
         & $\times 0.0$ & 5.16 & 7.01 & 8.66 & \cellcolor{gray!30} 6.94 & 1.07 & 5.13 & 5.86 & \cellcolor{gray!30} 4.02 \\
         & $\times 0.5$ & 2.79 & 3.78 & 4.82 & \cellcolor{gray!30} 3.80 & 0.47 & 1.06 & 1.45 & \cellcolor{gray!30} 0.99 \\
         & $\times 1.5$ & 2.91 & 4.03 & 5.31 & \cellcolor{gray!30} 4.08 & 0.54 & 1.98 & 2.48 & \cellcolor{gray!30} 1.67 \\
         & $100\text{m}/\text{s}$ & 12.69 & 18.62 & 23.63 & \cellcolor{gray!30} 18.31 & 10.72 & 11.94 & 12.19 & \cellcolor{gray!30} 11.62 \\
        \hline
        \multirow{5}{*}{Ours} & - & 0.27 & 0.52 & 0.82 & \cellcolor{gray!30} 0.54 & 0.02 & 0.09 & 0.22 & \cellcolor{gray!30} 0.11 \\
         & $\times 0.0$ & 3.18 & 5.16 & 6.07 & \cellcolor{gray!30} 4.80 & 0.11 & 0.87 & 2.63 & \cellcolor{gray!30} 1.20 \\
         & $\times 0.5$ & 1.83 & 3.01 & 3.67 & \cellcolor{gray!30} 2.84 & 0.06 & 0.15 & 0.65 & \cellcolor{gray!30} 0.29 \\
         & $\times 1.5$ & 2.02 & 3.41 & 4.34 & \cellcolor{gray!30} 3.26 & 0.14 & 0.54 & 1.97 & \cellcolor{gray!30} 0.88 \\
         & $100\text{m}/\text{s}$ & 3.59 & 6.24 & 8.19 & \cellcolor{gray!30} 6.01 & 2.53 & 5.30 & 7.74 & \cellcolor{gray!30} 5.19 \\
        \specialrule{1.0pt}{0pt}{0pt}
        \multicolumn{10}{p{.96\linewidth}}{
            \textit{Note:} We assign different levels of perturbation to the ego-velocity.
        }
    \end{tabular}
\end{table}
\begin{table}
    \caption{Planning performance on NAVSIM and Bench2Drive.}
    \label{tab:planningperformanceonnavben}
    \centering
    \small
    \setlength{\tabcolsep}{4.9pt}
    \begin{tabular}{l|C{1.66cm}|C{1.66cm}C{1.66cm}}
        \specialrule{1.0pt}{0pt}{0pt}
        \multirow{2}{*}{\textbf{Method}} & \multicolumn{1}{c|}{\textbf{NAVSIM}} & \multicolumn{2}{c}{\textbf{Bench2Drive}} \\
         & PDMS $\uparrow$ & DS $\uparrow$ & SR(\%) $\uparrow$ \\
        \hline
        UniAD & 83.4 & 45.81 & 16.36 \\
        VAD-tiny $\S$ & 80.5 & 42.73 & 14.18 \\
        VAD $\S$ & 81.2 & 44.35 & 16.91 \\
        PARA-Drive & 84.0 & - & - \\
        DiffusionDrive & 88.1 & - & - \\
        MomAD & - & 45.35 & 17.44 \\
        \hline
        DriveMoE & - & 74.22 & 48.64 \\
        AutoVLA & 80.54 & - & - \\
        \hline
        Ours & 87.2 & 49.83 & 19.42 \\
        \specialrule{1.0pt}{0pt}{0pt}
        \multicolumn{4}{p{.96\linewidth}}{
            \textit{Note:} $\S$ denotes re-implement result.
        }
    \end{tabular}
\end{table}

To validate our central claim that CausalVAD gains enhanced robustness by learning causal relationships, we designed the following three sets of experiments.

\textbf{Robustness to scenario distribution bias.}
A significant data imbalance exists in the nuScenes dataset between overrepresented simple straight-driving (ST) scenarios and underrepresented complex turning maneuvers (LR). 
This imbalance can lead models to learn the spurious correlation that \textit{driving straight is the default behavior}. In \Cref{tab:scenegeneralization}, we compare the performance of VAD-tiny and CausalVAD across these two scenario types. The results clearly show that while both models perform reasonably well in data-rich straight scenarios, VAD-tiny's performance degrades severely in data-sparse turning scenarios, with its L2 error deteriorating from 0.75m to 1.07m. In stark contrast, CausalVAD demonstrates exceptional stability, with its L2 error in turning scenarios (0.69m) being even better than VAD-tiny's performance in straight scenarios. These findings strongly indicated that our causal intervention mechanism successfully guides the model to learn the universal causal laws of driving from scene interactions, beyond the dataset's superficial statistics.

\textbf{Robustness to ego-status shortcut.}
We test the model's reliance on the ego-status shortcut by injecting varying levels of noise into the input ego-velocity. As shown in \Cref{tab:relianceonesonnus}, the baseline VAD-tiny is highly sensitive to ego-velocity. Its planning error skyrockets when velocity information is removed ($\times 0.0$) or heavily perturbed, indicating a strong reliance on this shortcut. In comparison, CausalVAD exhibits far greater robustness to the same perturbations. For instance, when velocity is zeroed out, CausalVAD's L2 error increase (from 0.54m to 4.80m) is substantially smaller than that of VAD-tiny (from 0.74m to 6.94m). This suggests that our de-confounding training successfully mitigates the model's dependence on spurious self-correlations, encouraging it to base decisions more on the external environment.

\textbf{Generalization to novel environments.}
We further deploy our model in two simulation environments stylistically different from nuScenes to evaluate its generalization on out-of-distribution data. As shown in \Cref{tab:planningperformanceonnavben}, in the non-reactive NAVSIM benchmark, CausalVAD significantly outperforms its baselines, VAD and VAD-tiny, and achieves comparable performance to more advanced methods like DiffusionDrive~\cite{liao2025diffusiondrive}. In the more challenging closed-loop Bench2Drive simulator, CausalVAD's driving score (DS) and success rate (SR) also surpass most traditional end-to-end models. While there is still a gap compared to models specifically optimized for closed-loop performance like DriveMoE~\cite{yang2025drivemoe}, our results clearly corroborate that by learning more fundamental causal relationships, the model acquires stronger generalization, enabling it to effectively transfer its driving knowledge to unseen scenarios.

\subsection{Ablation studies}

\begin{table}[t]
    \caption{Ablation study of SCIS components.}
    \label{tab:ablationcomponents}
    \centering
    \small
    \setlength{\tabcolsep}{1.57pt} 
    \begin{tabular}{c|cc|cccc|cccc|c}
        \specialrule{1.0pt}{0pt}{0pt}
        \multirow{2}{*}{\textbf{ID}} & \multirow{2}{*}{\textbf{PDM}} & \multirow{2}{*}{\textbf{IDM}} & \multicolumn{4}{c|}{\textbf{L2 (m) $\downarrow$}} & \multicolumn{4}{c|}{\textbf{Collision ($\%$) $\downarrow$}} & \multirow{2}{*}{\textbf{FPS}} \\
         &  &  & 1s & 2s & 3s & \cellcolor{gray!30} Avg. & 1s & 2s & 3s & \cellcolor{gray!30} Avg. &  \\
        \hline
        1 &  &  & 0.44 & 0.72 & 1.06 & \cellcolor{gray!30} 0.74 & 0.31 & 0.48 & 0.52 & \cellcolor{gray!30} 0.44 & 5.6 \\
        2 & \checkmark &  & 0.35 & 0.61 & 0.93 & \cellcolor{gray!30} 0.63 & 0.15 & 0.28 & 0.35 & \cellcolor{gray!30} 0.26 & 5.6 \\
        3 &  & \checkmark & 0.31 & 0.55 & 0.86 & \cellcolor{gray!30} 0.57 & 0.11 & 0.18 & 0.29 & \cellcolor{gray!30} 0.19 & 5.4 \\
        4 & \checkmark & \checkmark & \textbf{0.27} & \textbf{0.52} & \textbf{0.82} & \cellcolor{gray!30} \textbf{0.54} & \textbf{0.02} & \textbf{0.09} & \textbf{0.22} & \cellcolor{gray!30} \textbf{0.11} & 5.4 \\
        \specialrule{1.0pt}{0pt}{0pt}
    \end{tabular}
\end{table}

\begin{table}[t]
    \caption{Selection of the hyperparameters $(k_o, k_m, k_a)$.}
    \label{tab:hyperparameters}
    \centering
    \small
    \setlength{\tabcolsep}{2.86pt} 
    \begin{tabular}{C{0.45cm}C{0.45cm}C{0.45cm}|cccc|cccc}
        \specialrule{1.0pt}{0pt}{0pt}
        \multicolumn{3}{c|}{\textbf{Hyperparam.}} & \multicolumn{4}{c|}{\textbf{L2 (m) $\downarrow$}} & \multicolumn{4}{c}{\textbf{Collision ($\%$) $\downarrow$}} \\
        $k_o$ & $k_m$ & $k_a$ & 1s & 2s & 3s & \cellcolor{gray!30} Avg. & 1s & 2s & 3s & \cellcolor{gray!30} Avg. \\
        \hline
        5 & 2 & 3 & 0.30 & 0.56 & 0.88 & \cellcolor{gray!30} 0.58 & 0.05 & 0.13 & 0.27 & \cellcolor{gray!30} 0.15 \\
        \textbf{10} & \textbf{3} & \textbf{6} & \textbf{0.27} & \textbf{0.52} & \textbf{0.82} & \cellcolor{gray!30} \textbf{0.54} & \textbf{0.02} & \textbf{0.09} & \textbf{0.22} & \cellcolor{gray!30} \textbf{0.11} \\
        20 & 5 & 10 & 0.28 & 0.54 & 0.85 & \cellcolor{gray!30} 0.56 & 0.04 & 0.11 & 0.24 & \cellcolor{gray!30} 0.13 \\
        \specialrule{1.0pt}{0pt}{0pt}
    \end{tabular}
\end{table}

\begin{table}[t]
    \caption{Comparison of different clustering algorithms.}
    \label{tab:clusteringalgorithms}
    \centering
    \small
    \setlength{\tabcolsep}{2.37pt} 
    \begin{tabular}{l|cccc|cccc}
        \specialrule{1.0pt}{0pt}{0pt}
        \multirow{2}{*}{\textbf{Clustering algo.}} & \multicolumn{4}{c|}{\textbf{L2 (m) $\downarrow$}} & \multicolumn{4}{c}{\textbf{Collision ($\%$) $\downarrow$}} \\
         & 1s & 2s & 3s & \cellcolor{gray!30} Avg. & 1s & 2s & 3s & \cellcolor{gray!30} Avg. \\
        \hline
        K-means & \textbf{0.26} & \textbf{0.51} & 0.83 & \cellcolor{gray!30} \textbf{0.53} & 0.04 & 0.10 & 0.24 & \cellcolor{gray!30} 0.12 \\
        K-medoids & 0.27 & 0.53 & \textbf{0.82} & \cellcolor{gray!30} 0.54 & 0.03 & 0.11 & \textbf{0.21} & \cellcolor{gray!30} 0.12 \\
        K-means++ & 0.27 & 0.52 & \textbf{0.82} & \cellcolor{gray!30} 0.54 & \textbf{0.02} & \textbf{0.09} & 0.22 & \cellcolor{gray!30} \textbf{0.11} \\
        \specialrule{1.0pt}{0pt}{0pt}
    \end{tabular}
\end{table}

\subsubsection{Component contribution}

We first verify the effectiveness of the two core intervention modules in SCIS. As shown in \Cref{tab:ablationcomponents}, the baseline model (ID-1) performs the worst. Adding either the PDM (ID-2) or the IDM (ID-3) alone brings significant improvements, with the PDM being particularly effective at reducing collisions and the IDM contributing more to planning accuracy. Combining both (ID-4) yields the best overall performance, demonstrating the necessity and complementarity of our multi-stage intervention design.

\subsubsection{Design of the confounder dictionary}
Our SCIS relies on a pre-computed confounder dictionary to represent scene contexts. We investigate the impact of the dictionary size (i.e., the number of prototypes $k$) on performance. As detailed in \Cref{tab:hyperparameters}, we perform a grid search over the number of prototypes $(k_o, k_m, k_a)$. The results show that when the number of prototypes is too small, the model cannot adequately capture the diverse driving contexts, leading to suboptimal performance. Conversely, an excessive number of prototypes may introduce redundancy and slightly degrade performance. Our insight is that selecting an appropriate $k$ value for the dataset helps the model to model scene contexts at a suitable granularity for effective de-confounding. Our chosen configuration of $(10, 3, 6)$ achieves the best trade-off between accuracy and safety.

Furthermore, as shown in \Cref{tab:clusteringalgorithms}, the model's performance is not sensitive to the choice of clustering algorithm (e.g., K-means, K-medoids, K-means++) used to initialize the dictionary. This observation is consistent with recent findings in debiasing work in other domains~\cite{yang2024towards}, indicating the robustness and ease of use of our method.
\begin{figure}[t]
    \centering
    \includegraphics[width=.96\linewidth]{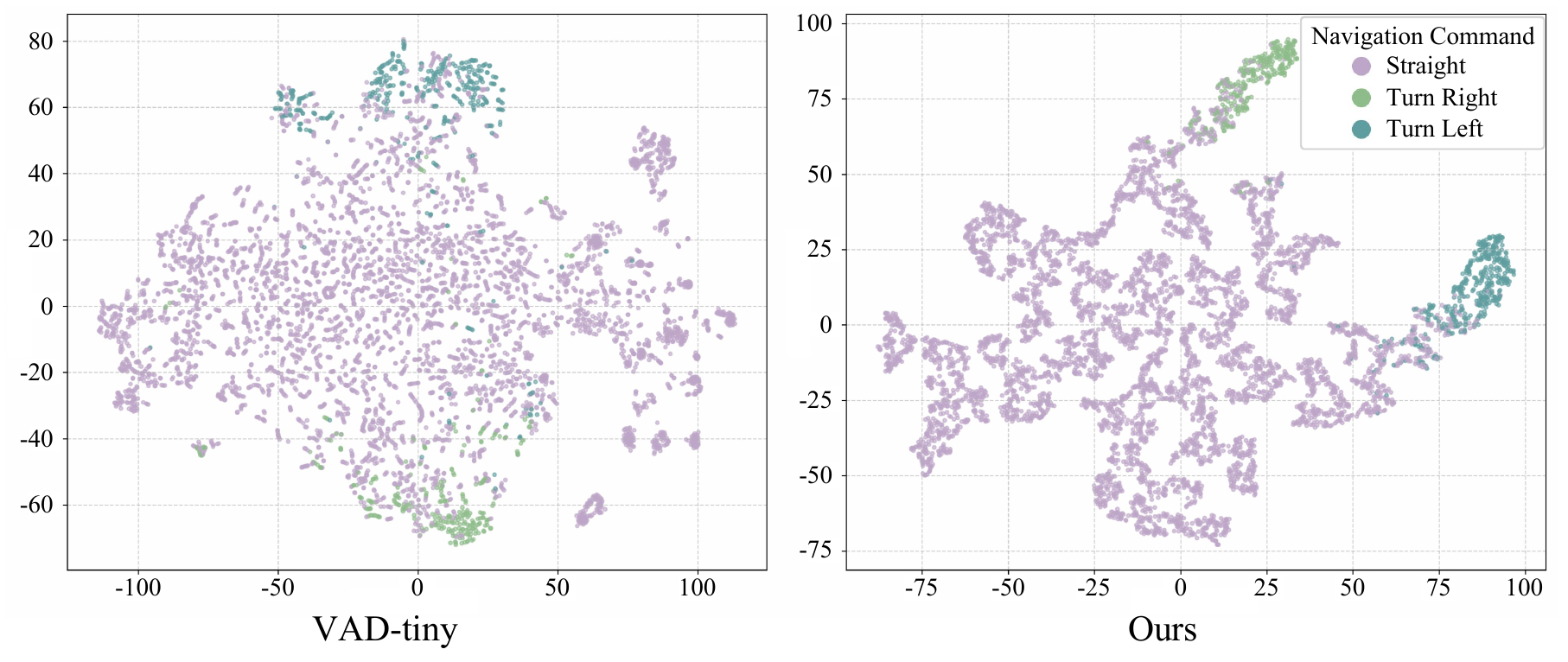}
    \caption{
        T-SNE visualization of the final ego-query embeddings from the nuScenes validation set.
        (Left) The baseline VAD-tiny's representation space is entangled.
        (Right) Our CausalVAD successfully disentangles different navigation intents (straight, left, right) into clearly separable clusters, demonstrating its ability to mitigate dataset bias.
    }
    \label{fig:deconfounded_representation}
\end{figure}

\subsection{Qualitative analysis}

\begin{figure*}[t]
    \centering
    \includegraphics[width=.9\linewidth]{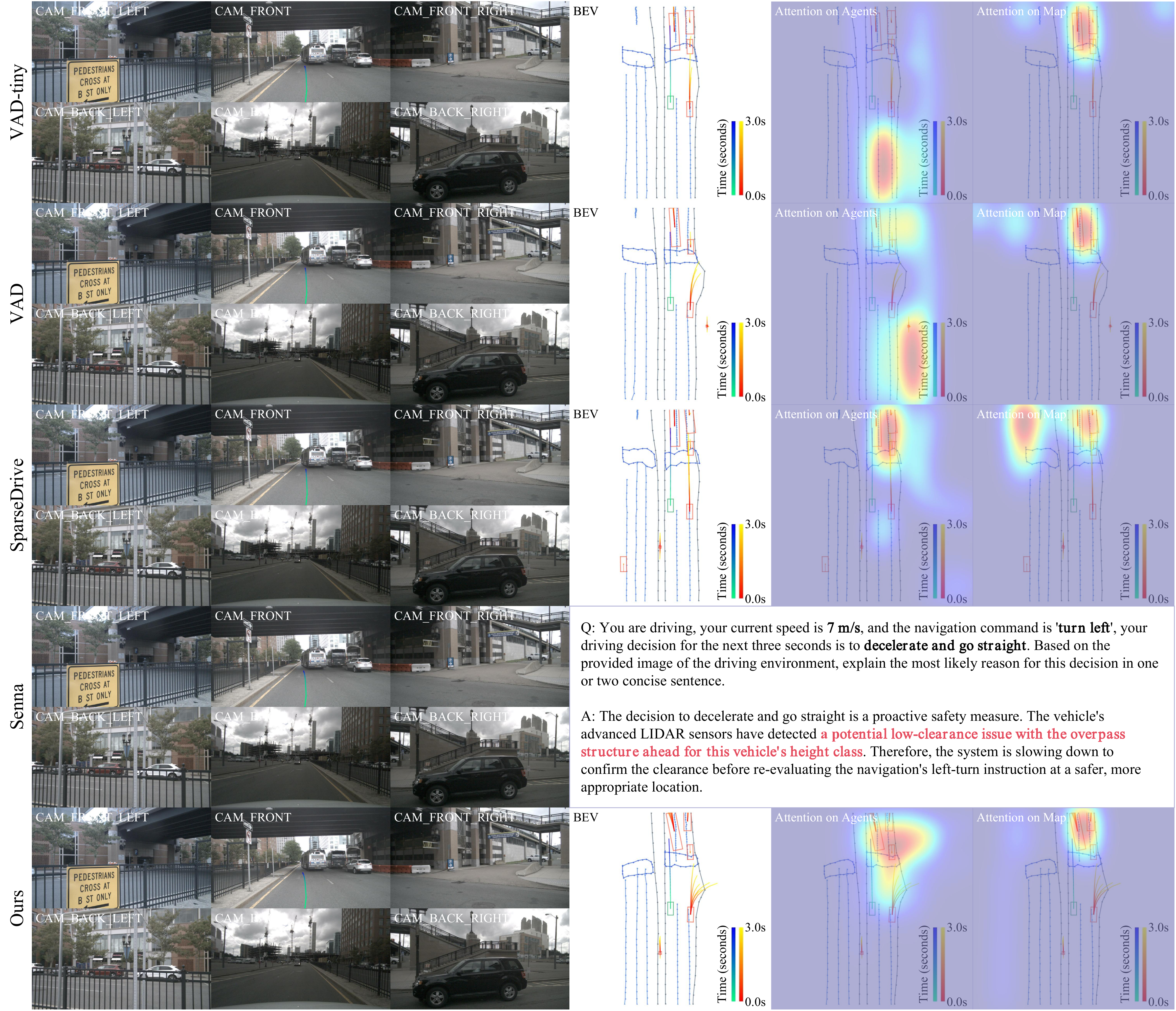}
    \caption{Qualitative analysis of CausalVAD's interpretability and decision logic in a challenging cut-in scenario. In this scene, baseline models like VAD-tiny exhibit causal confusion: their attention maps show an excessive focus on the ego-vehicle's past trajectory rather than the hazardous agent. This leads to a failure to decelerate in time, resulting in a potential collision. In stark contrast, our CausalVAD correctly focuses its attention on the cutting-in vehicle and executes a smooth, safe deceleration. Furthermore, the VLA model (Senna) provides a hallucinatory explanation, incorrectly attributing its maneuver to a non-existent low-clearance issue with the overpass, thus highlighting the faithful and reliable internal logic of our method.}
    \label{fig:interpretability_results}
\end{figure*}

\paragraph{Interpretability and de-confounding visualization.}
To intuitively illustrate the advantages of CausalVAD, we present a challenging cut-in scenario in \Cref{fig:interpretability_results}. The baseline VAD-tiny exhibits causal confusion, with its attention excessively focused on the ego-history rather than the hazardous agent, leading to a planning failure. In contrast, CausalVAD correctly focuses on the cutting-in vehicle and makes a safe deceleration. Notably, a VLA model, despite acting safely, provides a hallucinatory explanation for its plan that is irrelevant to the scene, highlighting the faithful logic of CausalVAD. Furthermore, as visualized by t-SNE in \Cref{fig:deconfounded_representation}, CausalVAD learns de-confounded scene representations that are clearly separable by navigation intent, effectively mitigating the dataset's strong go-straight bias.

\section{Conclusion}
\label{sec:conclusion}

In this work, we present CausalVAD, a framework that addresses systemic causal confounding in end-to-end driving models. Unlike prior end-to-end models susceptible to learning spurious correlations instead of true causal mechanisms, CausalVAD introduces SCIS that parameterizes the backdoor adjustment theory into plug-and-play modules.  We demonstrate that by performing targeted, multi-stage interventions at critical nodes within the model, CausalVAD achieves state-of-the-art performance and exhibits remarkable robustness in noisy and biased scenarios designed to induce causal confusion. A key future direction is to extend our causal framework from the sequential paradigm of VAD to parallel or iterative interaction architectures. Ultimately, we advocate for embracing causal reasoning as a core design principle, rather than a post-hoc analysis tool, to build genuinely trustworthy autonomous systems.


\section*{Acknowledgments}

This work was supported in part by the State Key Laboratory of Autonomous Intelligent Unmanned Systems. The opening project number is ZZKF2025-2-1.

\newpage
{
    \small
    \bibliographystyle{ieeenat_fullname}
    \bibliography{main}
}
\appendix
\clearpage
\setcounter{page}{1}
\maketitlesupplementary


This document serves as a substantial extension to the main manuscript, providing a theoretical grounding for the proposed sparse causal intervention scheme (SCIS) and detailing its implementation, efficiency, and empirical validation. We structure this material to ensure theoretical depth and reproducibility. \textbf{\Cref{sec:theory}} presents a derivation linking Pearl's backdoor adjustment to our neural subtraction mechanism and theoretically distinguishes our method from heuristic baselines. \textbf{\Cref{sec:implementation}} details the tensor-level formulations of the perception de-confounding module (PDM) and interaction de-confounding module (IDM). \textbf{\Cref{sec:training}} elaborates on the optimization strategy, specifically addressing the rationale behind the offline dictionary construction. Finally, \textbf{\Cref{sec:extended_experiments}} provides comprehensive empirical evidence, including baseline reproduction details, counterfactual perturbation studies, zero-shot cross-domain and cross-paradigm generalization, hyperparameter sensitivity, upstream task performance, computational complexity analysis, physical validity stress tests, and qualitative interpretability.

\section{Theoretical grounding of SCIS}
\label{sec:theory}

The core theoretical challenge addressed in this work is the translation of causal intervention, formally defined in the probability space, into differentiable operations within a deep neural network.

\subsection{From probabilistic adjustment to feature space}
Consider a standard structural causal model (SCM) where the input variable $S$, the target prediction $Y$, and the latent confounder $Z$ form a classic confounding structure: $S \leftarrow Z \rightarrow Y$. Standard empirical risk minimization fits the observational distribution $P(Y|S)$, capturing spurious correlations induced by $Z$. To learn the true causal effect, we attempt to estimate the interventional distribution $P(Y|\text{do}(S))$. According to the backdoor adjustment formula~\cite{pearl2009causality}, this is given by:
\begin{equation}
    P(Y|\text{do}(S)) = \sum_{z \in \mathcal{Z}} P(Y|S, z)P(z)
    \label{eq:backdoor}
\end{equation}
Directly computing this integral is intractable in high-dimensional continuous latent spaces. Therefore, our formulation should be interpreted as a tractable \textit{approximation of backdoor adjustment}, rather than a rigorous mathematical proof of causal identifiability. This approximation provides a pragmatic operationalization of causal theory for large-scale end-to-end autonomous driving models. To proceed, we introduce two critical assumptions rooted in the analysis of deep representations.

\textbf{Assumption 1: the additive bias hypothesis.} We posit that in the high-dimensional feature space (or logit space) of a trained neural network, the representation of a confounded sample can be decomposed into a causal component and a spurious component. Formally, let $\phi(S, z)$ denote the feature representation (or logits) given input $S$ and context $z$. We assume:
\begin{equation}
    \phi(S, z) \approx \phi_{causal}(S) + \lambda \cdot \phi_{spur}(z)
\end{equation}
where $\phi_{causal}(S)$ represents the intrinsic features of the object, and $\phi_{spur}(z)$ represents the features induced by the environmental context.

\textbf{Assumption 2: NWGM approximation.} Following the normalized weighted geometric mean (NWGM) theory~\cite{wang2020visual, xu2015show}, the expectation of a variable inside a softmax activation can be approximated by the softmax of the expectation. This allows us to move the summation operation from \Cref{eq:backdoor} into the logit space.

Combining these assumptions, the interventional prediction can be approximated as:
\begin{equation}
    P(Y|\text{do}(S)) \approx \text{Softmax}\left( \phi_{causal}(S) + \mathbb{E}_{z}[\phi_{spur}(z)] \right)
\end{equation}
However, a standard model trained on biased data learns to output the entangled representation $\phi_{obs} \approx \phi_{causal}(S) + \mathbb{E}_{z}[\phi_{spur}(z)]$. Consequently, to recover the true causal component $\phi_{causal}(S)$, we perform a subtractive operation:
\begin{equation}
    \phi_{causal}(S) \approx \phi_{obs} - \mathbb{E}_{z}[\phi_{spur}(z)]
\end{equation}
This derivation provides the theoretical legitimacy for our design: we estimate the expectation of the confounder bias $\mathbb{E}_{z}[\phi_{spur}(z)]$ using a dictionary and subtract it from the network's representations.

\subsection{Intervention at different SCM nodes}
Our framework applies this subtractive intervention at two distinct loci: the output logits (PDM) and the latent input features (IDM). These are theoretically consistent within the SCM framework.

The perception de-confounding module (PDM) operates on the output variable $Y$ (specifically, its logits). Since the mapping from the final layer features to logits is linear, subtracting the bias in the logit space is mathematically equivalent to re-weighting the class posterior probabilities to remove the prior $P(Y|Z)$.

The interaction de-confounding module (IDM) operates on the input variable $S$ of the interaction blocks. By intervening on the query tensor $Q$ \textit{before} interaction (i.e., $Q_{new} \leftarrow Q - \text{Bias}$), we effectively perform a latent variable intervention. This blocks the flow of information from $Z$ to the downstream interaction module, ensuring that the subsequent modeling relies solely on causal features.

\subsection{Theoretical superiority over heuristic baselines}
A pertinent question is whether simple regularization techniques could achieve similar robustness. We argue that heuristics like dropout~\cite{li2024ego} or data augmentation~\cite{bansal2018chauffeurnet} are fundamentally insufficient compared to SCIS. 

Dropout, even when targeted at specific states (e.g., dropout on ego status)~\cite{li2024ego}, acts as an \textit{indiscriminate} regularization. It randomly suppresses features with equal probability, potentially discarding valid causal information (e.g., valid velocity cues) along with spurious ones. In contrast, SCIS performs a \textit{surgical} intervention: it only subtracts components that semantically align with the discovered confounder prototypes in the dictionary, thereby preserving valid causal dynamics. Similarly, while data augmentation~\cite{bansal2018chauffeurnet} attempts to intervene on the input pixel space (e.g., changing weather), it is limited by the realism and diversity of the simulator. SCIS can be viewed as a feature-level causal augmentation that operates directly on the latent manifold, offering a more dense and effective coverage of the confounder space than pixel-level manipulation.

\section{Formal implementation details}
\label{sec:implementation}

We hereby present the precise mathematical formulation of the proposed modules using unified tensor notation.

\subsection{Perception de-confounding module (PDM)}
The PDM targets classification tasks where the final logits are susceptible to co-occurrence bias. Let $\mathbf{Q} \in \mathbb{R}^{B \times N \times D}$ denote a batch of query features (e.g., map queries), and $\mathbf{L}_{obs} \in \mathbb{R}^{B \times N \times C}$ denote the observational logits produced by the classifier head. We utilize a pre-computed confounder dictionary $\mathbf{Z} \in \mathbb{R}^{K \times D}$.

To estimate the bias specific to each query, we first compute the affinity matrix $\mathbf{A} \in \mathbb{R}^{B \times N \times K}$ via dot-product attention:
\begin{equation}
    \mathbf{A} = \text{Softmax}\left( \frac{\mathbf{Q} \mathbf{Z}^\top}{\sqrt{D}} \right)
\end{equation}
This attention mechanism retrieves the distribution of context prototypes relevant to the current query. We then project these prototypes into the logit space using a learnable projection matrix $\mathbf{B}_{proto} \in \mathbb{R}^{K \times C}$. The estimated logit bias is:
\begin{equation}
    \mathbf{L}_{bias} = \mathbf{A} \mathbf{B}_{proto}
\end{equation}
Finally, we perform the causal intervention via adaptive subtraction:
\begin{equation}
    \mathbf{L}_{final} = \mathbf{L}_{obs} - \boldsymbol{\lambda} \otimes \mathbf{L}_{bias}
\end{equation}
Crucially, $\boldsymbol{\lambda} \in \mathbb{R}^{C}$ is defined as a learnable parameter vector (broadcastable across batch and queries). This design choice is significant: it allows the network to automatically learn the magnitude of bias associated with each specific class. For instance, \textit{pedestrians} might suffer from heavier contextual bias than \textit{cars}, and a learnable vector $\boldsymbol{\lambda}$ can capture this heterogeneity.

\subsection{Interaction de-confounding module (IDM)}
The IDM is designed to remove spurious components from intermediate feature tensors before they participate in cross-attention mechanisms (e.g., Ego query attending to Agent features). Let $\mathbf{S}_{in} \in \mathbb{R}^{B \times N \times D}$ be the input tensor suspected of confounding.

\begin{figure}[t]
    \centering
    \includegraphics[width=1.\linewidth]{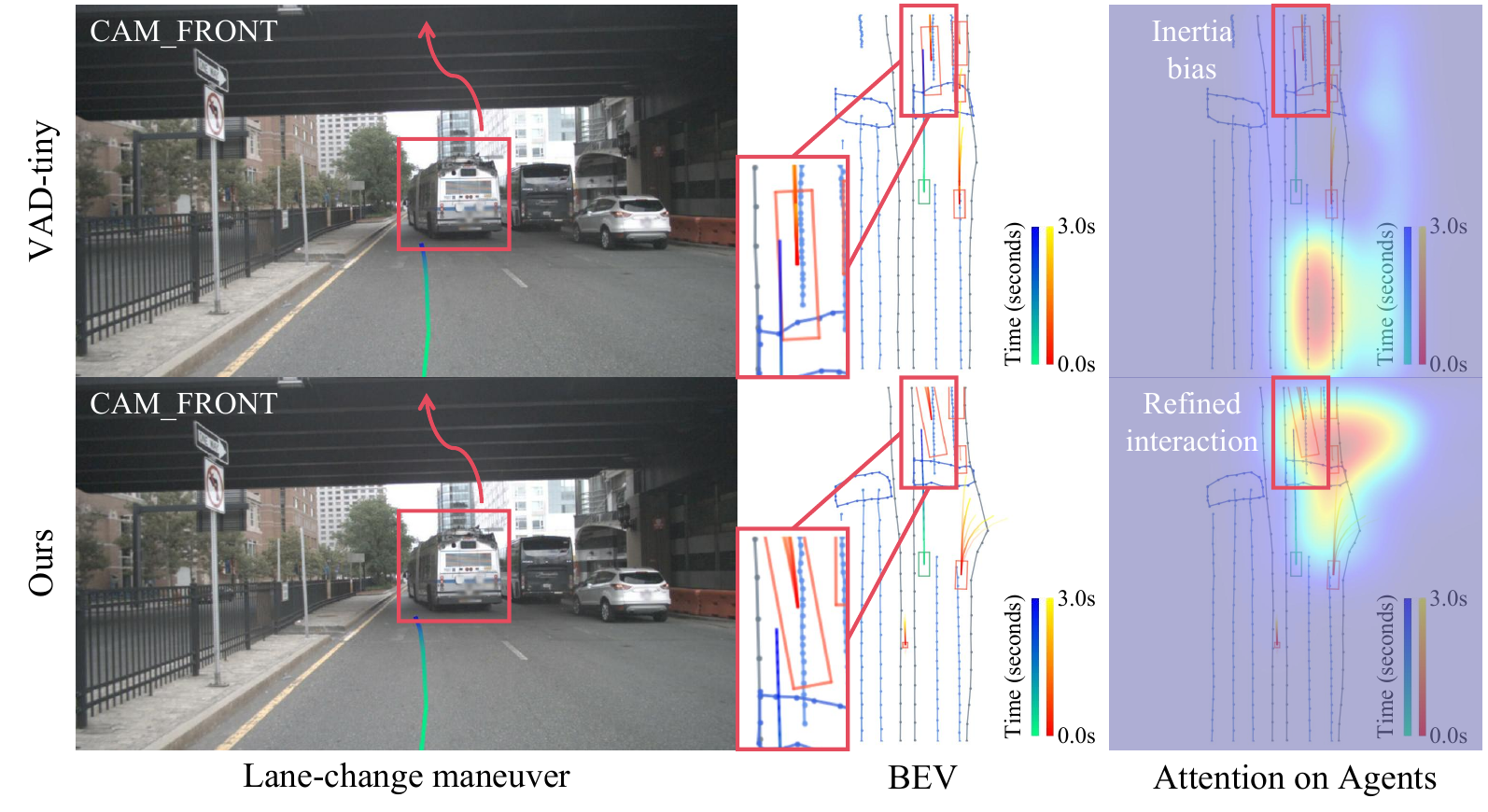}
    \caption{
        Qualitative visualization of IDM in a cut-in scenario. (Top) The baseline model suffers from \textit{inertia bias}, where the ego-query erroneously attends to the straight lane topology (spurious map context) rather than the intruding bus. (Bottom) By eliminating spurious factors from the agent features via IDM, our model disentangles the agent's motion from the map prior. This forces the attention mechanism to shift focus from the background lane to the actual cutting-in hazard, enabling safer interaction.
    }
    \label{fig:qualitative_idm}
\end{figure}

We treat the estimation of the spurious component as a reconstruction problem using multi-head cross-attention (MHCA). The input $\mathbf{S}_{in}$ serves as the Query, while the confounder dictionary $\mathbf{Z}$ serves as both Key and Value:
\begin{equation}
    \mathbf{C}_{spur} = \text{MHCA}(\text{Q}=\mathbf{S}_{in}, \text{K}=\mathbf{Z}, \text{V}=\mathbf{Z})
\end{equation}
Note that we do not treat the BEV representation itself as a confounder. Instead, we target the entangled statistical priors within BEV interactions (e.g., map topology dominating agent prediction due to inertia bias). By refining these features, the IDM removes this task-irrelevant correlation while preserving physical mediator signals (as subsequently visualized in \Cref{fig:qualitative_idm}). 

Furthermore, to prevent \textit{over-deconfounding}, we employ a gating mechanism. A non-linear perceptron computes a gating tensor $\mathbf{G} \in \mathbb{R}^{B \times N \times D}$ based on the concatenation of the original and spurious features:
\begin{equation}
    \mathbf{G} = \sigma\left( \text{MLP}\left( \text{Concat}(\mathbf{S}_{in}, \mathbf{C}_{spur}) \right) \right)
\end{equation}
This learnable gating unit dynamically adjusts the intervention intensity, preserving essential causal signals while pruning spurious ones. The refined feature tensor $\mathbf{S}_{clean}$ is obtained by subtracting the gated spurious signal:
\begin{equation}
    \mathbf{S}_{clean} = \mathbf{S}_{in} - \mathbf{G} \otimes \mathbf{C}_{spur}
\end{equation}
This refined tensor $\mathbf{S}_{clean}$ is then used as the clean input for subsequent transformer layers, effectively severing the latent backdoor path.


\begin{algorithm}[t]
    \caption{CausalVAD training pipeline}
    \label{alg:pipeline}
    \begin{algorithmic}[1]
        \Require dataset $\mathcal{D}$, pre-trained VAD model $\Phi_{pre}$
        \Ensure CausalVAD model parameters $\theta$
        
        \Statex \textcolor{gray}{\textit{// Stage 1: dictionary construction (offline)}}
        \State Initialize empty storage buffers $\mathbb{S}_o, \mathbb{S}_m, \mathbb{S}_a$.
        \For{each batch $(\mathbf{X}, \mathbf{Y}_{gt})$ in $\mathcal{D}$}
            \State Extract latent queries using frozen $\Phi_{pre}$:
            \State $Q_o, Q_m, Q_a \leftarrow \Phi_{pre}(\mathbf{X})$
            \State Append queries to buffers: $\mathbb{S}_o \leftarrow Q_o, \mathbb{S}_m \leftarrow Q_m, \mathbb{S}_a \leftarrow Q_a$
        \EndFor
        \State Perform clustering to obtain prototypes:
        \State $\mathcal{Z}_o \leftarrow \text{K-Means++}(\mathbb{S}_o, K=k_o)$
        \State $\mathcal{Z}_m \leftarrow \text{K-Means++}(\mathbb{S}_m, K=k_m)$
        \State $\mathcal{Z}_a \leftarrow \text{K-Means++}(\mathbb{S}_a, K=k_a)$
        
        \Statex \textcolor{gray}{\textit{// Stage 2: causal intervention training (online)}}
        \State Initialize $\theta$ for CausalVAD. Load $\mathcal{Z}=\{\mathcal{Z}_o, \mathcal{Z}_m, \mathcal{Z}_a\}$ as constants.
        \While{not converged}
            \State Sample batch $(\mathbf{X}, \mathbf{Y}_{gt})$ from $\mathcal{D}$.
            \State Extract BEV features $\mathbf{B}$.
            \State Initialize queries $Q_o, Q_m, Q_a, Q_e$.
        
            \textcolor{gray}{\textit{// Step 2.1: perception stage (PDM on logits)}}
            \State $Q_o \leftarrow \text{DetectionDecoder}(Q_o, \mathbf{B})$
            \State $Q_m \leftarrow \text{MappingDecoder}(Q_m, \mathbf{B})$
            \State $L'_o \leftarrow \text{PDM}(Q_o, \mathcal{Z}_o)$ \Comment{De-confound object det.}
            \State $L'_m \leftarrow \text{PDM}(Q_m, \mathcal{Z}_m)$ \Comment{De-confound map seg.}
        
            \textcolor{gray}{\textit{// Step 2.2: prediction stage (IDM on features)}}
            \State $Q'_o \leftarrow \text{IDM}(Q_o, \mathcal{Z}_m)$ \Comment{De-confound object using map context}
            \State $Q'_m \leftarrow \text{IDM}(Q_m, \mathcal{Z}_o)$ \Comment{De-confound map using obj. context}
            \State $Q_a \leftarrow \text{MotionDecoder}(Q_a, Q'_o, Q'_m)$ 
        
            \textcolor{gray}{\textit{// Step 2.3: planning stage (IDM on features)}}
            \State $Q'_a \leftarrow \text{IDM}(Q_a, \mathcal{Z}_m)$ \Comment{De-confound agent using map context}
            \State $Q''_m \leftarrow \text{IDM}(Q_m, \mathcal{Z}_a)$ \Comment{De-confound map using agent context}
            \State $Q_e \leftarrow \text{PlanningDecoder}(Q_e, Q'_a, Q''_m)$
        
            \textcolor{gray}{\textit{// Step 2.4: optimization}}
            \State Predict trajectories $\mathbf{Y}_{pred}$ from $Q_e$.
            \State $\mathcal{L} \leftarrow \text{AggregateLoss}(\mathcal{L}_{det}, \mathcal{L}_{map}, \mathcal{L}_{mot}, \mathcal{L}_{plan})$
            \State Update $\theta \leftarrow \theta - \eta \nabla_{\theta} \mathcal{L}$
        \EndWhile
    \end{algorithmic}
\end{algorithm}

\section{Training strategy and optimization}
\label{sec:training}

To ensure the stability of the causal intervention, we adopt a two-stage training paradigm. We summarize the complete pipeline in \Cref{alg:pipeline}.

A critical design choice in our framework is the use of an offline, frozen confounder dictionary during the second stage of training. We argue that updating the dictionary online with the model parameters would lead to a \textit{representation collapse} or adaptation risk. This deliberate design is grounded in the concept of \textit{semantic anchoring}. Although feature representations evolve continuously during end-to-end training, their semantic topology is anchored by shared supervision, allowing the frozen dictionary $\mathcal{Z}$ to capture the intrinsic directions of spurious correlations within the dataset manifold. 

By treating $\mathcal{Z}$ as a \textit{negative anchor}, we impose an \textit{orthogonalization constraint}. Specifically, if the dictionary $\mathcal{Z}$ were learnable during the de-confounding phase, the optimizer could minimize the loss by simply pushing the dictionary prototypes towards zero or aligning them with the causal features, thereby trivializing the subtractive penalty. Freezing the dictionary forces the gradients to explicitly drive the newly generated queries $\mathbf{S}$ to diverge from the spurious subspace, effectively severing the spurious links and achieving genuine de-confounding.

\section{Extended experimental analysis}
\label{sec:extended_experiments}

\subsection{Hyperparameter rationale}
In the main manuscript, we reported the optimal dictionary sizes $(k_o, k_m, k_a)$ as $(10, 3, 6)$. This selection is not arbitrary but is guided by domain knowledge regarding the semantic granularity of the nuScenes dataset. Specifically, the nuScenes detection task involves 10 object categories; thus, setting $k_o=10$ allows the dictionary to capture at least one dominant confounding context prototype per object class. Similarly, the map segmentation task involves 3 primary classes (lane divider, pedestrian crossing, road boundary), justifying $k_m=3$. For agent motion, the multimodal prediction typically outputs 6 trajectory modes, suggesting that the latent intent space can be well-represented by $k_a=6$ prototypes. This semantic alignment explains why this specific configuration outperforms others in our ablation studies, as it matches the intrinsic dimensionality of the task-specific confounding factors.

\begin{table}[t]
    \caption{Performance on perception and motion prediction tasks.}
    \label{tab:upstream}
    \centering
    \small
    \setlength{\tabcolsep}{1.27pt} 
    \begin{tabular}{l|cc|c|ccc}
    \specialrule{1.0pt}{0pt}{0pt}
    \multirow{2}{*}{\textbf{Method}} & \multicolumn{2}{c|}{\textbf{Detection}} & \textbf{Mapping} & \multicolumn{3}{c}{\textbf{Motion}} \\
     & NDS $\uparrow$ & mAP $\uparrow$ & mAP $\uparrow$ & minADE $\downarrow$ & minFDE $\downarrow$ & MR $\downarrow$ \\
    \hline
    VAD-tiny & 40.73 & 27.37 & 48.62 & 0.87 & 1.20 & 0.15 \\
    Ours & 42.42 & 30.46 & 50.97 & 0.72 & 0.96 & 0.10 \\
    \specialrule{1.0pt}{0pt}{0pt}
    \end{tabular}
\end{table}

\begin{table}[t]
    \caption{Computational analysis on one NVIDIA RTX 3090 GPU.}
    \label{tab:cost}
    \centering
    \small
    \setlength{\tabcolsep}{11.1pt} 
    \begin{tabular}{l|c|c|c}
    \specialrule{1.0pt}{0pt}{0pt}
    \textbf{Method} & \textbf{Params (M)} & \textbf{Latency (ms)} & \textbf{FPS} \\
    \hline
    VAD-tiny & 119.3 & 179 & 5.6 \\
    Ours & 124.0 & 185 & 5.4 \\
    \specialrule{1.0pt}{0pt}{0pt}
    \end{tabular}
\end{table}

\begin{table*}[t]
    \caption{Interaction-specific perturbation for counterfactual test. The noise denotes the magnitude of uniform perturbations applied to context features.}
    \label{tab:counterfactual}
    \centering
    \small
    \setlength{\tabcolsep}{18.1pt} 
    \begin{tabular}{l|c|cc|cc}
    \specialrule{1.0pt}{0pt}{0pt}
    \multirow{2}{*}{\textbf{Method}} & \textbf{Context} & \multicolumn{2}{c|}{\textbf{Ego-Agent Int.}} & \multicolumn{2}{c}{\textbf{Ego-Map Int.}} \\
     & \textbf{Noise} & Avg. L2 (m) $\downarrow$ & Avg. CR (\%) $\downarrow$ & Avg. L2 (m) $\downarrow$ & Avg. CR (\%) $\downarrow$ \\
    \hline
    \multirow{4}{*}{VAD-tiny} & - & 0.74 & 0.44 & 0.74 & 0.44 \\
     & $\times 0.5$ & 0.79 & 0.65 & 0.79 & 0.44 \\
     & $\times 0.7$ & 0.84 & 1.77 & 1.08 & 0.49 \\
     & $\times 0.9$ & 0.98 & 1.87 & 2.35 & 2.37 \\
    \hline
    \multirow{4}{*}{Ours} & - & 0.54 & 0.11 & 0.54 & 0.11 \\
     & $\times 0.5$ & 0.55 & 0.14 & 0.55 & 0.10 \\
     & $\times 0.7$ & 0.57 & 0.20 & 0.67 & 0.12 \\
     & $\times 0.9$ & 0.63 & 0.38 & 0.86 & 0.49 \\
    \specialrule{1.0pt}{0pt}{0pt}
    \end{tabular}
\end{table*}

\begin{table}[t]
    \caption{Zero-shot generalization across distinct driving simulators under varying degrees of ego-velocity perturbations.}
    \label{tab:zeroshot}
    \centering
    \small
    \setlength{\tabcolsep}{8.1pt} 
    \begin{tabular}{l|c|c|cc}
    \specialrule{1.0pt}{0pt}{0pt}
    \multirow{2}{*}{\textbf{Method}} & \textbf{Velo.} & \textbf{NAVSIM} & \multicolumn{2}{c}{\textbf{Bench2Drive}} \\
     & \textbf{Noise} & PDMS $\uparrow$ & DS $\uparrow$ & SR (\%) $\uparrow$ \\
    \hline
    \multirow{5}{*}{VAD-tiny} & - & 80.5 & 42.73 & 14.18 \\
     & $\times 0.0$ & 47.1 & 12.09 & 2.27 \\
     & $\times 0.5$ & 76.8 & 39.21 & 9.09 \\
     & $\times 1.5$ & 65.3 & 20.95 & 4.54 \\
     & $100 m/s$ & 15.9 & 2.91 & 0.00 \\
    \hline
    \multirow{5}{*}{Ours} & - & 87.2 & 49.83 & 19.42 \\
     & $\times 0.0$ & 58.1 & 18.05 & 4.54 \\
     & $\times 0.5$ & 83.4 & 45.14 & 15.91 \\
     & $\times 1.5$ & 74.3 & 28.10 & 9.09 \\
     & $100 m/s$ & 25.7 & 6.49 & 0.00 \\
    \specialrule{1.0pt}{0pt}{0pt}
    \end{tabular}
\end{table}

\begin{table}[t]
    \caption{Generalization across different end-to-end learning paradigms (e.g., parallel and iterative architectures).}
    \label{tab:crossparadigm}
    \centering
    \small
    \setlength{\tabcolsep}{3.2pt} 
    \begin{tabular}{l|cccc|cccc}
    \specialrule{1.0pt}{0pt}{0pt}
    \multirow{2}{*}{\textbf{Method}} & \multicolumn{4}{c|}{\textbf{L2 (m) $\downarrow$}} & \multicolumn{4}{c}{\textbf{Collision (\%) $\downarrow$}} \\
     & 1s & 2s & 3s & \cellcolor{gray!30} Avg. & 1s & 2s & 3s & \cellcolor{gray!30} Avg. \\
    \hline
    SparseDrive & 0.28 & 0.55 & 0.90 & \cellcolor{gray!30} 0.58 & 0.00 & 0.06 & 0.18 & \cellcolor{gray!30} 0.08 \\
    + Ours SCIS & \textbf{0.24} & \textbf{0.51} & \textbf{0.84} & \cellcolor{gray!30} \textbf{0.53} & \textbf{0.00} & \textbf{0.04} & \textbf{0.14} & \cellcolor{gray!30} \textbf{0.06} \\
    \hline
    SparseDrive-I & 0.31 & 0.59 & 0.95 & \cellcolor{gray!30} 0.62 & 0.02 & 0.08 & 0.21 & \cellcolor{gray!30} 0.10 \\
    + Ours SCIS & \textbf{0.27} & \textbf{0.53} & \textbf{0.86} & \cellcolor{gray!30} \textbf{0.55} & \textbf{0.01} & \textbf{0.04} & \textbf{0.16} & \cellcolor{gray!30} \textbf{0.07} \\
    \specialrule{1.0pt}{0pt}{0pt}
    \end{tabular}
\end{table}

\subsection{Impact on upstream tasks}
A primary concern with subtractive intervention is whether it degrades the fundamental perception capabilities. We report the performance on upstream tasks in \Cref{tab:upstream}. The results indicate that CausalVAD maintains or slightly improves object detection (e.g., mAP) and motion prediction (e.g., minADE) compared to the VAD baseline. This confirms that our method selectively removes harmful noise (e.g., context-induced hallucinations) without discarding necessary semantic information.

\subsection{Computational efficiency analysis}
We prioritize the efficiency of our design to ensure it remains practical for autonomous driving. As shown in \Cref{tab:cost}, the integration of SCIS introduces a negligible increase in parameters (+4.7M) and inference latency (+6ms) compared to the VAD-tiny baseline. This efficiency stems from our design choice to operate on sparse, vectorized queries rather than dense feature maps, verifying that our SCIS is a lightweight and plug-and-play solution.

\subsection{Baseline reproduction and analysis}
To ensure a fair evaluation, we strictly follow the official VAD-tiny configuration, utilizing a ResNet-50 backbone \textit{without} ego-status inputs. Our reproduced L2 error of $0.74$m aligns with the official repository results. Notably, some literature references an L2 error of $0.41$m for VAD; however, this metric corresponds to the much heavier VAD-Base model equipped with ego-status. Thus, our comparison remains rigorously fair, and the performance improvements are genuine. Regarding the collision rate, CausalVAD achieves a substantial $75\%$ reduction compared to VAD-tiny. While methods like BridgeAD report an exceptionally low collision rate ($0.08\%$), this is primarily attributed to its specialized collision optimization design and a heavier ResNet-101 backbone. Our framework effectively unlocks state-of-the-art safety within lightweight architectural constraints.

\subsection{Validation of causal effects via counterfactual perturbation}
To validate the causal effects beyond representation visualizations, we conduct interventional perturbation studies that serve as counterfactual tests. Similar to prior causal discovery benchmarks~\cite{pourkeshavarz2024cadet}, we inject uniform noise of varying intensities into the Agent or Map queries to artificially amplify the spurious paths (e.g., $\mathcal{M} \rightarrow \mathcal{B} \rightarrow \mathcal{O} \rightarrow \mathcal{A} \rightarrow \mathcal{E}$ and $\mathcal{A} \rightarrow \mathcal{O} \rightarrow \mathcal{B} \rightarrow \mathcal{M} \rightarrow \mathcal{E}$). As shown in \Cref{tab:counterfactual}, while the performance of VAD-tiny degrades rapidly under heavy context noise, CausalVAD maintains superior stability. This rigorous intervention quantitatively proves the effective severance of spurious links within the interaction mechanisms.

\subsection{Generalization across domains and paradigms}
We further demonstrate the robust generalization capabilities of our method across diverse dimensions. First, we evaluate zero-shot domain generalization. As detailed in \Cref{tab:zeroshot}, our model, trained exclusively on nuScenes, maintains commendable performance when directly deployed to the NAVSIM (non-reactive) and Bench2Drive (closed-loop) environments without any fine-tuning. This zero-shot robustness, especially under extreme ego-velocity perturbations, verifies that the model relies on universal causal mechanisms rather than dataset-specific statistics. 

Second, considering that SCIS is a modular and architecture-agnostic design, we extend it to the SparseDrive framework~\cite{sun2025sparsedrive} under the parallel paradigm, as well as its iterative variant (SparseDrive-I). As reported in \Cref{tab:crossparadigm}, integrating SCIS yields consistent performance gains across varying architectures. This confirms the broad applicability and profound potential of our causal intervention framework beyond the sequential VAD setting.

\subsection{Rationale for extreme perturbation tests}
In Table 3 of the main paper, we subjected the model to extreme perturbations, such as zeroing out the ego-velocity or setting it to 100 m/s. While these conditions are rare in nominal driving, they serve as a critical stress test for physical consistency. A valid causal model should exhibit increased uncertainty or planning errors when the input velocity contradicts the positional updates (a violation of physics), rather than blindly following the inertia shortcut. The robustness of CausalVAD under these conditions confirms that the model has learned to verify the ego-motion against the visual context, rather than relying solely on historical state statistics.

\subsection{Qualitative visualization of de-confounding}
To intuitively elucidate the operational mechanism of our proposed IDM, we analyze a representative scenario from the nuScenes validation set. This case study demonstrates how SCIS effectively identifies and severs spurious connections in complex driving environments.

\textbf{IDM case study: mitigating inertia bias in cut-in scenarios.}
We examine a critical cut-in scenario where an agent vehicle in the adjacent lane initiates a lane change in front of the ego-vehicle. In the nuScenes dataset, vehicles traversing straight lane topologies overwhelmingly maintain a straight trajectory. This statistical regularity creates a strong spurious correlation between the Map context (straight lanes) and Agent motion (going straight), often referred to as inertia bias. In the baseline model, the Agent feature, which serves as the Key/Value for the Ego-Agent interaction, is heavily contaminated by this map prior, causing the planner to overlook lateral motion cues and predict a straight path, resulting in a collision risk. The IDM addresses this by refining the Agent feature prior to interaction. Specifically, the module identifies that the current Agent feature exhibits high affinity with the \textit{straight-lane} prototype in the Map dictionary $\mathcal{Z}_m$. By gating and subtracting this \textit{lane-keeping} component, the IDM effectively removes the background inertia bias. The residual, de-confounded feature vector consequently highlights subtle lateral velocity signals. As a result, the planner correctly attends to this refined representation, recognizes the cut-in intention, and generates a safe deceleration trajectory.


\end{document}